\documentclass[sigconf]{acmart}

\AtBeginDocument{%
  }

\setcopyright{acmcopyright}
\copyrightyear{2023}
\acmYear{2023}
\acmDOI{XXXXXXX.XXXXXXX}

\acmConference[WSDM '23]{Proceedings of the Sixteenth ACM International Conference on Web Search and Data Mining}{Feb 27-- March 03,
  2023}{In-person Conference with Virtual Elements, Singapore}
\acmPrice{15.00}
\acmISBN{978-1-4503-XXXX-X/18/06}

\begin{document}

\title{PiggyBack: Pretrained Visual Question Answering Environment \\for Backing up Non-deep Learning Professionals}

\author{
Zhihao Zhang$^{1}$\authornotemark{*},
Siwen Luo$^{1}$\authornotemark{*},
Junyi Chen$^{1}$,
Sijia Lai$^{1}$,\\
Siqu Long$^{1}$,
Hyunsuk Chung$^{3}$,
Soyeon Caren Han$^{1, 2}$\authornotemark{$\dag$}
}


\affiliation{%
  \institution{$^{1}$The University of Sydney, $^{2}$The University of Western Australia, $^{3}$FortifyEdge}
  \state{$^{1, 3}$New South Wales, $^{2}$Western Australia}
  \country{Australia}
}

\email{{zhihao.zhang1, siwen.luo, caren.han}@sydney.edu.au, {slai3926, slon6753}@uni.sydney.edu.au} \email{david.chung@fortifyedge.com, junyi.chen.215@gmail.com}

\renewcommand{\shortauthors}{Zhang and Luo, et al.}
\renewcommand*{\thefootnote}{\fnsymbol{footnote}}

\begin{abstract}
\footnotetext[1]{Both authors contributed equally to this research.} 
\footnotetext[2]{Corresponding author.}
\renewcommand*{\thefootnote}{\arabic{footnote}}

We propose a PiggyBack, a Visual Question Answering platform that allows users to apply the state-of-the-art visual-language pretrained models easily. The PiggyBack supports the full stack of visual question answering tasks, specifically data processing, model fine-tuning, and result visualisation. We integrate visual-language models, pretrained by HuggingFace, an open-source API platform of deep learning technologies; however, it cannot be runnable without programming skills or deep learning understanding. Hence, our PiggyBack supports an easy-to-use browser-based user interface with several deep learning visual language pretrained models for general users and domain experts. The PiggyBack includes the following benefits: Free availability under the MIT License, Portability due to web-based and thus runs on almost any platform, A comprehensive data creation and processing technique, and ease of use on deep learning-based visual language pretrained models. The demo video is available on YouTube and can be found at \url{https://youtu.be/iz44RZ1lF4s}.
\end{abstract}


\maketitle

\section{Introduction}
\renewcommand*{\thefootnote}{\arabic{footnote}}
Visual Question Answering (VQA)~\cite{antol2015vqa} is a Vision-and-Language task that requires answering natural language questions by referring to relevant regions of a given image. This task has proved its practical assistance in various real-world applications, including automatic medical diagnosis~\cite{vu2020question,ren2020cgmvqa}, visual-impaired people guidance~\cite{ren2020cgmvqa}, education assistance \cite{he2017educational} and customer advertising improvement~\cite{zhou2020recommending}. To achieve acceptable performances, task-specific models require large-scale datasets to learn the visual and textual features sufficiently~\cite{long2022vision}. However, domain-related datasets could be low-resource due to the collection difficulties and expensiveness, especially in the medical domain. For example, the largest radiology dataset SLAKE~\cite{SLAKE} only contains 14K image-question pairs. The Vision-and-Language Pre-trained Models (VLPMs) become helpful in this case. VLPMs are pretrained on huge image-text dataset collections to learn the generic representations of the visual and textual alignment~\cite{long2022gradual}, which various downstream vision-and-language tasks can then use. Recently, several large VLPMs~\cite{li2019visualbert, vlbert, lxmert} have been proposed and have proved their state-of-the-art performances on the downstream VQA task. These large VLPMs empower the merit of transfer learning and can be smoothly adapted to different domains by fine-tuning on small-scale datasets while maintaining competitive performances. Therefore, VLPMs can be an excellent solution to the lack of domain-related data. VLPMs have become popular among deep learning researchers, and many open-source tools and APIs are publicly released. Nevertheless, VLPMs are not vastly applied in industrial domains. This is because such implementation requires solid deep learning and programming skills and thus is challenging for non-deep learning experts.

\textbf{Contribution.} With this in mind, we propose PiggyBack, a deep learning web-based interactive VQA platform, to help field experts such as physicians, educators and commercial analysts. Our PiggyBack is mainly for those who lack deep learning expertise or programming skills and easily apply VLPMs on VQA tasks with their dataset. More precisely, Piggyback provides two pre-trained models, and users can freely choose and train one of the models over their training data by interacting with its user interface. It also supports model evaluation directly on users' testing sets with numerous image-question pairs. It enhances the evaluation results with interpretability by visualising the relevant regions for question-answering on the image. Such interpretation would help users build confidence in the model's decision, especially for critical fields. 




\textbf{Comparison.} To the best of our knowledge, PiggyBack is the first web-based deep-learning platform that provides a user-friendly interface for non-deep learning users. It allows the users to train VQA models with their datasets by utilising VLPMs in the manner of transfer learning (also known as fine-tuning) and testing the model with their testing datasets. Some of the existing VQA platforms are not based on VLPMs, such as Simple Baseline for VQA\footnote[1]{\url{http://visualqa.csail.mit.edu/}} and Explainable VQA\footnote[2]{\url{https://lrpserver.hhi.fraunhofer.de/visual-question-answering/}}, which cannot provide the benefit of the generalised pre-trained model. Other VQA platforms only focus on testing the models' performance by evaluating the single image-question pair, such as CloudCV\footnote[3]{\url{http://visualqa.csail.mit.edu/}}, ViLT VQA\footnote[4]{\url{https://huggingface.co/spaces/nielsr/vilt-vqa}} and OFA-VQA\footnote[5]{\url{https://huggingface.co/spaces/OFA-Sys/OFA-vqa}}, which cannot be trained towards users’ datasets. Furthermore, none of these platforms combines and simplifies the training and testing procedures to provide the VLPMs' capability for other field experts.

\section{PiggyBack System}
\begin{figure}[h]
  \centering
  \includegraphics[width=0.478\textwidth]{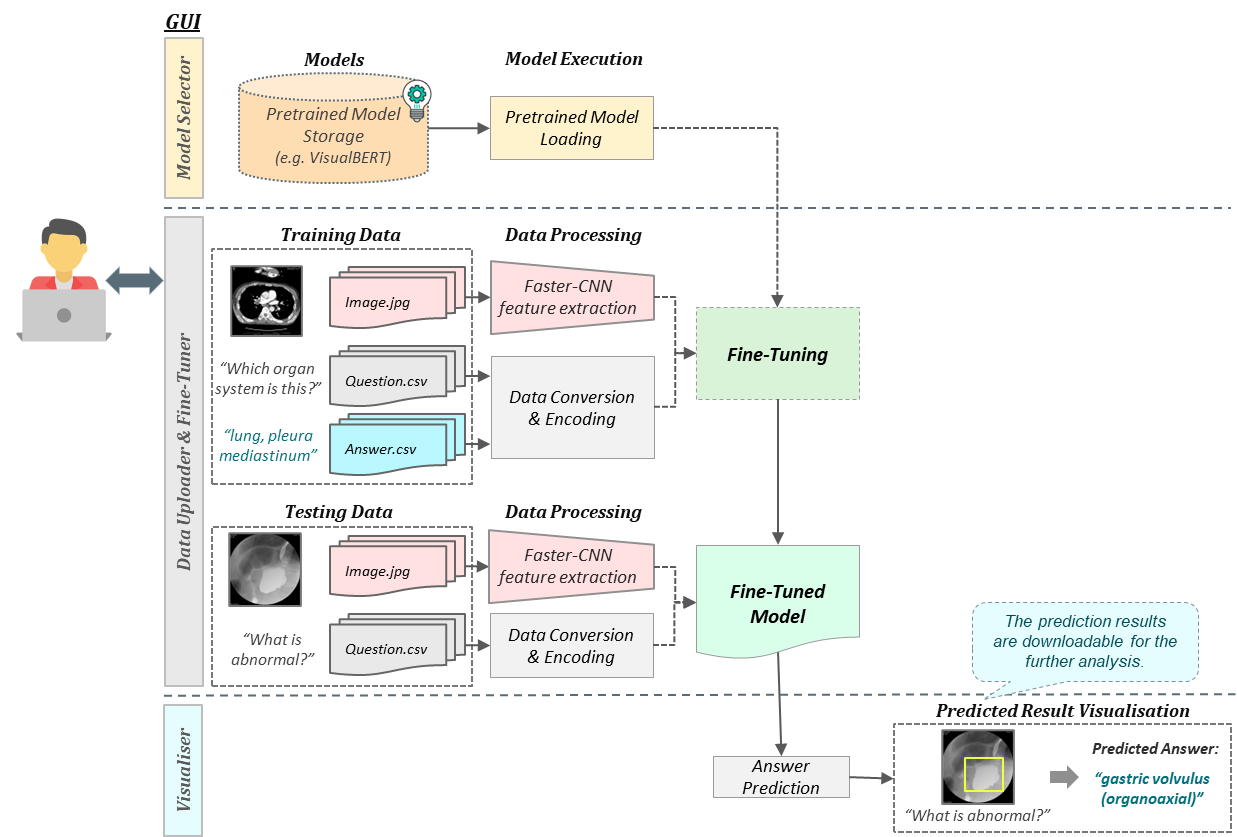}
  \caption{PiggyBack platform framework, which consists of four main components: 1)Data Uploader, 2)Model Selector, 3)Fine-Tuner, and 4)Visualiser.}
  \label{fig:flow}
\end{figure}

PiggyBack integrates the VLPMs implemented by HuggingFace Transformer~\cite{wolf-etal-2020-transformers} while keeping all the coding away from users behind the well-designed browser-based Graphic User Interface (GUI). Therefore, we designed both the backend and front-end of the system to standardise the workflow scenario for VQA tasks, so any non-deep learning/non-programming professionals can utilise PiggBack effortlessly. Its design flow is shown in Figure~\ref{fig:flow}, which illustrates the front-end components in Sec.4. PiggyBack’s backend and front-end are described in the following sections.

\section{Backend Architecture}
The system backend is built upon the Flask framework~\cite{grinberg2018flask}. Since it contains no database abstraction layer, the input data is handled by Python and saved in the server's local environment. The backend includes four components that cover all necessary procedures in model fine-tuning and evaluation.

\subsection{Data Preparation} 

PiggyBack simplifies the users’ data preparation by providing the automatic data cleaning process and asking for simple dataset formats, which can be easily prepared. It includes a zip file including all the images and a CSV file containing all the questions and their ground-truth answers for the associated images. As the backend models require the input data following the VQAv2’s one-question ten-answers format~\cite{VQA}, we developed a module in our system to clean the imperfect data in users’ uploaded CSV. The cleaning steps include: 1) auto-fill the 10 answers when users did not provide enough answers; 2) remove the duplicated image question pairs accidentally provided by the users or the questions that have no valid image id; 3) remove the images that exceed the required size. The cleaned CSV is then transferred into JSON format, which can be directly loaded into different models. Visual features are extracted from images by a feature extraction module and saved into JSON format. This stand-alone module is containerised by Docker~\cite{merkel2014docker} and implements the Bottom-Up, and Top-Down Attention model~\cite{Anderson2017up-down}. Such data cleaning processes are all wrapped in the backend, which leaves users an easy and simplified experience in their data preparation step.

\subsection{Model Structure} 
Inspired by V-Doc~\cite{Ding_2022_CVPR}, our system includes two state-of-the-art pre-trained models: VisualBERT and LXMERT, which offer the users an opportunity to conduct the performance comparison of models with different structures and enable them to choose the model that suits their data the best. VisualBERT~\cite{li2019visualbert} encodes the visual embedding as the sum of bonding region features, segment embedding and position embedding. In the meantime, it encodes the textual embedding following the BERT format, including token embeddings, segment embeddings and position embedding. A single Transformer structure is proposed in VisualBERT, which uses visual and textural embeddings to discover alignments between vision and language. VisualBERT is pre-trained with Masked Language Modeling with the Image Task and Sentence-image Prediction Task, and it can be fine-tuned with VQA datasets. LXMERT~\cite{lxmert} directly takes a sequence of objects from images as the visual inputs and a sequence of words from sentences as the linguistic inputs. There are three Transformer encoders inside LXMERT, which separately encode image object features, question features and cross-modality interactions. LXMERT is pre-trained with five tasks, including Masked Cross-Modality Langage Model, RoI-Feature Regression, Detected-Label Classification Cross-Modality Matching and Image Question Answering, and it can be fine-tuned for VQA downstream task. Both pre-trained models are built upon the HuggingFace deep-learning API~\cite{wolf-etal-2020-transformers}, and have proved to be an outstanding performance on the VQA tasks.

\subsection{Fine Tuning} 
Once the model is selected, PiggyBack loads the pre-trained model and finds the answer space from the preprocessed data. Then it feeds the data into the data loader and launches the fine-tuning process with the specific answer space on the pre-trained model. When the fine-tuning operation finishes, the fine-tuned model will be packed into a loadable file, which can be imported for evaluation. All the fine-tuning procedures are handled by the backend, so there is no deep-learning knowledge required from the users.

\subsection{Visualised Evaluation}\label{sec:backendVE}
PiggyBack allows the fine-tuned model to evaluate with numerous image-question pairs and delivers the predictions in a single CSV file. Apart from the predicted answers, PiggyBack embeds a visualisation module, which enhances the model interpretability by annotating the important object regions in the images according to their attention scores. Attention scores have long been used as a feature-based local interpretation method for deep neural networks. Both VisualBERT and LXMERT utilise the Transformer structure with the multi-head self-attention mechanism~\cite{vaswani2017attention}. For the visual component, the attention mechanism assigns attention weights for each region of the input images. The region with higher attention weights is naturally considered more critical to the model's outputs~\cite{han2020victr}. We sum up the attention weights across all heads for all transformer layers as the final attention score for each object region and visualise the top 5 object regions with the highest attention scores and annotate them with their bounding boxes. Regions with higher attention scores are marked in a darker colour.

\section{Front-end Interactive Web Page}
PiggyBack provides an interactive web front-end that is built upon the Bootstrap\footnote[1]{\url{https://getbootstrap.com/}} framework. We established three pages to cover the four components in the backend. The home page includes Data Uploader, Model Selector and Fine-tuner, which introduces a straightforward interface for input datasets uploading, model choosing and training operating. The progress page shows the fine-tuning progress. The evaluation page includes Visualiser, which illustrates the models' performance to the users after fine-tuning. Those web pages aim to guide the users in completing the fine-tuning and evaluation process. 

\subsection{Data Uploader} 
\begin{figure}[ht!]
  \centering
  \includegraphics[width=0.85\linewidth]{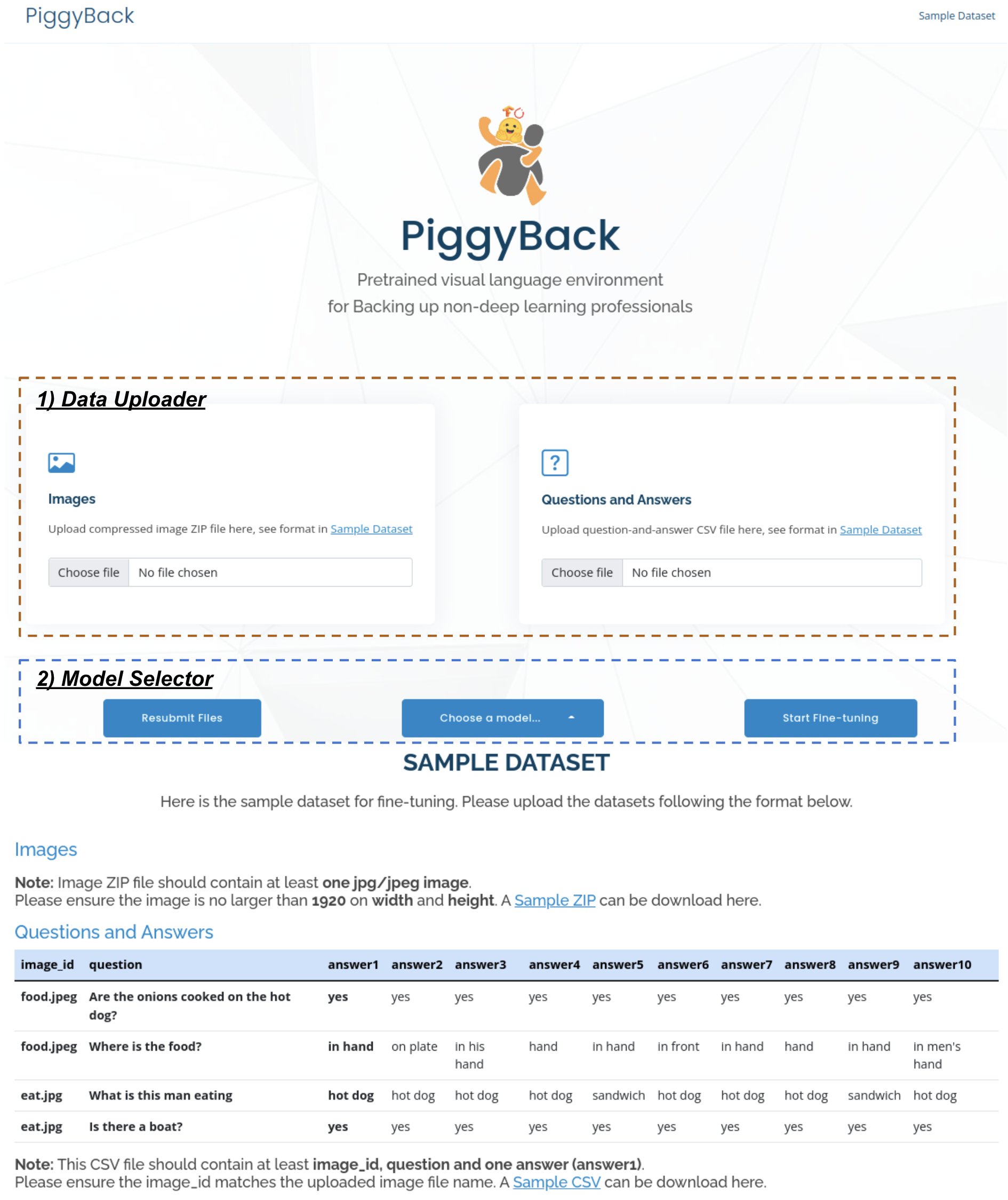}
  \caption{This figure shows the Preview of Data Uploader and Model Selector on the front-end interface in our PiggyBack.}
  \label{fig:uploader}
\end{figure}

Our system landing page presents the GUI of the Data Uploader for collecting the user's training dataset, which is shown in Figure~\ref{fig:uploader}. Under the "Images" and "Questions and Answers" sections, the users only need to upload a compressed ZIP folder that includes all images as well as a CSV file that contains all the questions and answers with the corresponding image id. There are two constraints placed in the uploaded dataset due to the prerequisite of the VLPMs: 1) the input image's width and height should be within 1920 pixels; 2) the input question, answer and image id should be legitimate to form one piece of the data. To help the users comprehensively understand the data format, we provide the illustrations on the data uploading page and the \textit{"Sample Dataset"} files that can be downloaded and even modified by users with their data. 

The interactive web page can provide prompt feedback to the users when the image data is sent into the Data Uploader: 
1) if there is no valid image in the uploaded folder, a red Error banner will show up, and it requires a new image folder from the users; 
2) if there are some oversized images in the uploaded folder, a yellow Warning banner will show up, and the users can choose to fine-tune without those images or resubmit the image folder after modification; 
3) if all the images meet the constraint, a green Success banner will show up, and the users can choose to fine-tune with current images or resubmit the image folder. The input question-and-answer data in CSV is preprocessed in the system backend.

\subsection{Model Selector and Fine-tuner}

After uploading the dataset successfully, the users can choose either VisualBERT or LXMERT by simply selecting if from the \textit{``Choose a model''} drop-down menu in the interface. Once the users click \textit{``Start fine-tuning''} with the chosen pre-trained model, all the processed data features will be passed to the loaded pre-trained model for the fine-tuning process. Meanwhile, a progress bar appears on the web page, indicating the completion of the fine-tuning step. If the users neglect the model selection but click \textit{``Start fine-tuning''}, a red Error banner will show up asking to select the pre-trained model, and the system will hold the fine-tuning process till a pre-trained model has been selected. 

\subsection{Visualiser}

\begin{figure}[ht!]
  \centering
  \includegraphics[width=0.85\linewidth]{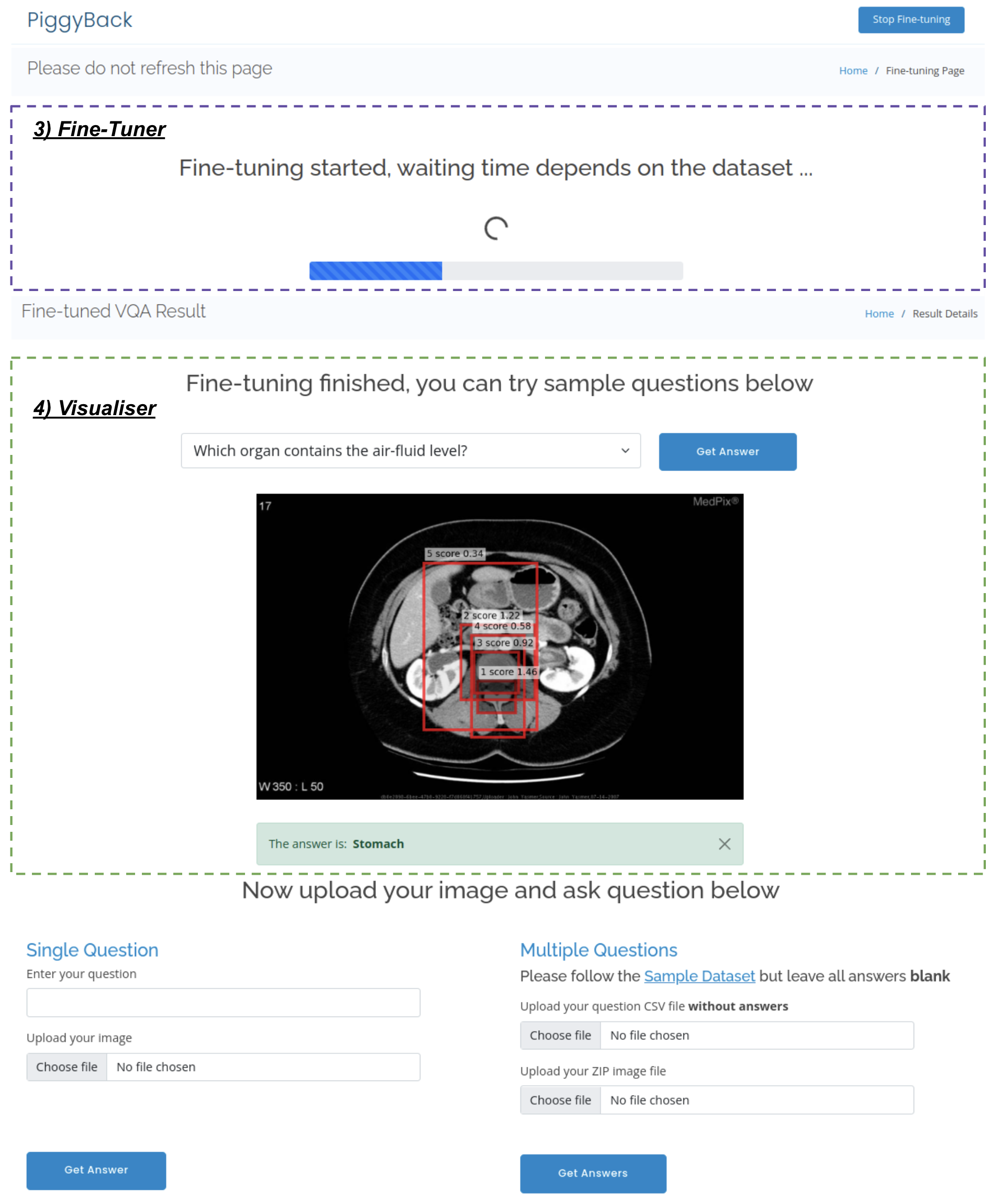}
  \caption{This figure shows the Preview of Fine-tuner and Visualiser on the front-end interface in our PiggyBack.}
  \label{fig:visualiser}
\end{figure}

Once fine-tuning finishes, users will be automatically redirected to the evaluation page, which incorporates the Visualiser.  As shown in Figure~\ref{fig:visualiser}, we provide three different scenarios for users to test the fine-tuned model:

\textbf{Sample evaluation.} This evaluation section equips with a sample case at the top half of the page. Both visual and textual outputs are shown directly on the page, so the users can have a quick glance at the model's performance and understand the visualised outputs from the PiggyBack. We put a radiology image and medical-related questions as an example; the users can select different questions in the drop-down menu and click \textit{"Get Answer"} to see the predicted answers. Furthermore, the Visualiser annotates the top five regions in the image based on the significance calculated by the fine-tuned model. The insights of the significance calculation are introduced in Sec.\ref{sec:backendVE}.

\textbf{Single evaluation.} PiggyBack provides a testing GUI for the users, which allows them to upload a single image and ask a question about it. As shown in Figure~\ref{fig:visualiser}, this section is at the bottom left of the evaluation page. The system shows the uploaded image's preview on the page, which ensures that they type in the relevant question. Similar to the sample evaluation above, a predicted answer and its corresponding annotated image will appear on the page upon clicking \textit{``Get Answer''}.

\textbf{Multiple evaluation.} Apart from single evaluation, PiggyBack is capable of multiple image-question pairs evaluation, which is more practical in the real-world scenario. This section is at the bottom right of the evaluation page. The required format of the multiple evaluation data is similar to the training data; the only difference is that the answers are not required in the testing CSV. In this evaluation GUI, the instruction of the CSV modification and the hyperlink to the previous sample dataset is provided for the users, which helps them with the testing data preparation. We designed a simple checking mechanism, which shows a red error message when there is no valid image or question entry in the dataset. After uploading the testing data, the users can click \textit{``Get Answers''} to get model predictions, and a green banner will show up with the download links for both annotated image ZIP and answer CSV. The annotated images in the ZIP are renamed with their questions, which helps the user easily combine the model predictions with the corresponding images.

\section{Conclusion}

PiggyBack is a web-based vision-and-language modelling platform that aims to support non-deep learning users utilizing the SOTA VLPMs for VQA problems in their specific domains. The PiggyBack system provides a user-friendly interface that simplifies all the data uploading, model fine-tuning and evaluation with only a few clicks. Meanwhile, it accompanies the results with a straightforward interpretation to help users better understand the model's decision. 




\bibliographystyle{ACM-Reference-Format}
\bibliography{sample-base}










\end{document}